\documentclass[11pt,a4paper]{article}
\usepackage[hyperref]{emnlp2018}
\usepackage{array}
\usepackage{times}
\usepackage{latexsym}
\usepackage{pdflscape}
\usepackage{booktabs}
\usepackage{longtable}
\usepackage[german]{babel}
\usepackage{rotating}
\usepackage{url}
\usepackage{setspace}
\usepackage{hanging}
\usepackage{amsmath}

\aclfinalcopy 

\title{Fine-grained evaluation of German-English \\ Machine Translation based on a Test Suite}

\author{Vivien Macketanz, Eleftherios Avramidis, Aljoscha Burchardt, Hans Uszkoreit \\
  German Research Center for Artificial Intelligence (DFKI), Berlin, Germany \\
  {\tt firstname.lastname@dfki.de} \\}

\date{}

\begin{document}
\maketitle

\begin{abstract}
We present an analysis of 16 state-of-the-art MT systems on German-English based on a linguistically-motivated test suite. 
The test suite has been devised manually by a team of language professionals
in order to cover a broad variety of linguistic phenomena that MT often fails to translate properly.
It contains 5,000 test sentences covering 106 linguistic phenomena in 14 
categories, with an increased focus on verb tenses, aspects and moods.
The MT outputs are evaluated in a semi-automatic way through regular expressions that focus only on the part of the sentence that is relevant to each phenomenon. 
Through our analysis, we are able to compare systems based on their performance
on these categories.
Additionally, we reveal strengths and weaknesses of particular systems and we
identify grammatical phenomena where the overall performance of MT is relatively low.

\end{abstract}

\section{Introduction}

The evaluation of Machine Translation (MT) has mostly relied on methods that
produce a numerical judgment on the correctness of a test set. These methods are
either based on the human perception of the correctness of the MT output
\cite{callisonburch-EtAl:2007:WMT}, or on automatic metrics that compare the MT
output with the reference translation \cite{papineni02:bleu,snover06}.
In both cases, the evaluation is performed on a test-set containing articles or
small documents that are assumed to be a random representative sample of
texts in this domain.
Moreover, this kind of evaluation aims at producing average scores that express
a generic sense of correctness for the entire test set and compare the
performance of several MT systems.

Although this approach has been proven valuable for the MT development and the
assessment of new methods and configurations, it has been suggested that a more
fine-grained evaluation, associated with linguistic phenomena, may lead in a
better understanding of the errors, but also of the efforts required to improve
the systems \cite{pub8367}.
This is done through the use of test suites, which are carefully devised corpora,
whose test sentences include the phenomena that need to be tested. In this paper
we present the fine-grained evaluation results of 16 state-of-the-art MT systems
on German-English, based on a test suite focusing on 106 German grammatical
phenomena with a focus on verb-related phenomena. 

\section{Related Work} 
The use of test suites in the evaluation of NLP applications
\citep{balkan1995test} and MT systems in particular \citep{King1990,Way1991}
has been proposed already in the 1990's.
For instance, test suites were employed to evaluate state-of-the-art rule-based
systems \citep{heid1991some}.
The idea of using test suites for MT evaluation was revived recently with the
emergence of Neural MT (NMT) as the produced translations reached significantly
better levels of quality, leading to a need for more fine-grained qualitative
observations.
Recent works include test suites that focus on the evaluation of particular
linguistic phenomena \cite[e.g.
pronoun translation;][]{Guillou2016} or more generic test suites that aim at
comparing different MT technologies \citep{Isabelle2017,Burchardt2017} and
Quality Estimation methods \cite{Avramidis2018}.
The previously presented papers differ in the amount of phenomena and the
language pairs they cover.

This paper extends the work presented in \newcite{Burchardt2017} by including
more test sentences and better coverage of phenomena.
In contrast to that work, which applied the test suite in order to compare 3
different types of MT systems (rule-based, phrase-based and NMT), the evaluation in the publication at hand has been applied on 16 state-of-the-art systems whose majority follows the
NMT methods.

\section{Method}

This test suite is a manually devised test set, aiming to investigate the MT
performance against a wide range of linguistic phenomena or other qualitative
requirements (e.g. punctuation).

It contains a set of sentences in the source language, written or chosen by a
team of linguists and professional translators with the aim to cover as many
linguistic phenomena as possible, and particularly the ones that MT often fails
to translate properly.
Each sentence of the test suite serves as a paradigm for investigating only one
particular phenomenon.
Given the test sentences, the evaluation tests the ability of the MT systems to
properly translate the associated phenomena.
The phenomena are organized in categories (e.g. although each verb tense is
tested separately with the respective test sentences, the results for all tenses
are aggregated in the broader category of verb tense/aspect/mood).

Our test suite contains about 5,000 test sentences, covering 106 phenomena
organized in 14 categories.
For each phenomenon at least 20 test sentences were devised to allow better
generalizations about the capabilities of the MT systems.
With 88\%, the majority of the test suite covers verb phenomena, but other categories, such
as negation, long distance dependencies, valency or multi-word expressions are
 included as well.
A full list of the phenomena and their categories can be seen in
Table~\ref{tab:categorization}. An example list of test sentences with correct and incorrect  translations is available on GitHub\footnote{\url{https://github.com/DFKI-NLP/TQ_AutoTest}}.

\subsection{Construction of the Test Suite}

\begin{figure*}
\includegraphics[width=\textwidth]{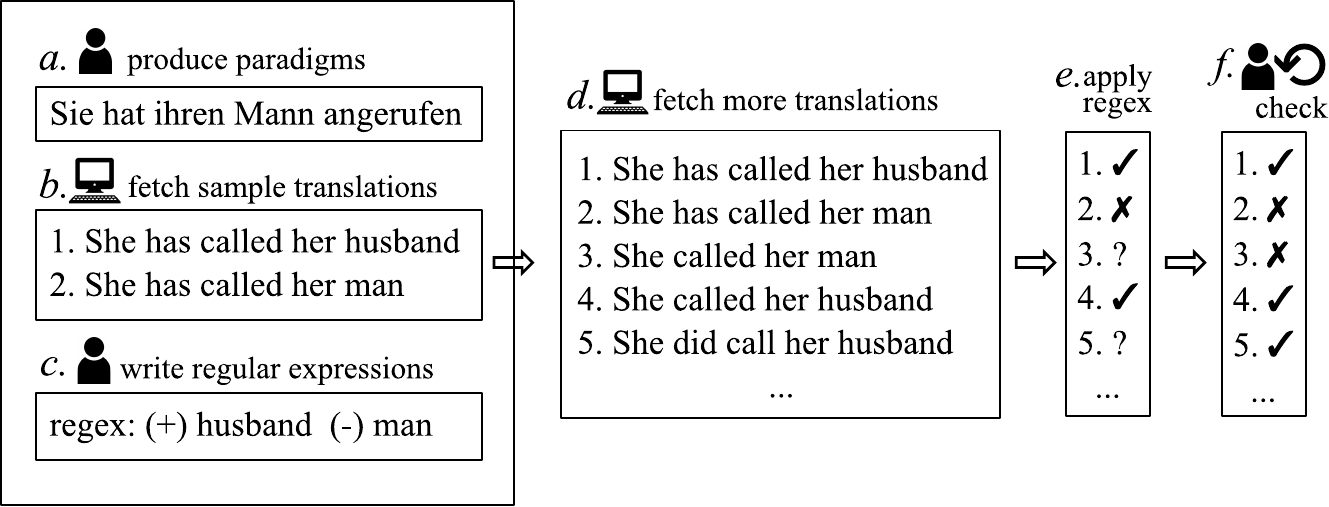}
\caption{Example of the preparation and application of the test suite for one
test sentence}
\label{fig:construction}
\end{figure*}

The test suite was constructed in a way that allows a semi-automatic evaluation
method, in order to assist the efficient evaluation of many translation systems.
A simplified sketch of the test suite construction is shown in
Figure~\ref{fig:construction}.
First (Figure~\ref{fig:construction}, stage~a), the linguist choses or writes
the test sentences in the source language with the help of translators. 
The test sentences are manually written or chosen, based on whether their
translation has demonstrated or is suspected to demonstrate MT errors of the
respective error categories.
Test sentences are selected from various parallel corpora or drawn from existing
resources, such as the TSNLP Grammar Test Suite \cite{Lehmann1996} and online
lists of typical translation errors.
Then (stage~b) the test sentences are passed as an input to the some sample MT
systems and their translations are fetched.

Based on the output of the sample MT systems and the types of the errors, the
linguist devises a set of hand-crafted regular expressions (stage~c) while  the translator ensures the correctness of the expressions.
The regular expressions are used to automatically check if the output correctly
translates the part of the sentence that is related to the phenomenon under
inspection.
There are regular expressions that match correct translations (positive) as
well as regular expressions that match incorrect translations (negative).  

\subsection{Application of the Test Suite}

During the evaluation phase, the test sentences are given to several translation
systems and their outputs are acquired (stage~d).
The regular expressions are applied to the MT outputs (stage~e) to automatically
check whether the MT outputs translate the particular phenomenon properly.
An MT output is marked as correct (\emph{pass}), if it matches a positive
regular expression.
Similarly, it is marked as incorrect (\emph{fail}), if it matches a negative
regular expression.
In cases where the MT output does not match either a positive or a negative
regular expression
, the automatic evaluation flags an
uncertain decision (\emph{warning}).
Then, the results of the automatic annotation are given to a linguist or a
translator who manually checks the warnings (stage~f) and optionally refines the
regular expressions in order to cover similar future cases.
It is also possible to add full sentences as valid translations, instead of
regular expressions.
In this way, the test suite grows constantly, whereas the required manual effort
is reduced over time.

Finally, for every system we calculate the phenomenon-specific translation
accuracy:

\begin{equation*} 
 \textrm{accuracy} =  \frac{\textrm{correct\;translations}}{\textrm{sum\;of\;test\;sentences}} 
\end{equation*} 

\noindent The translation accuracy per phenomenon is given by the number of the test sentences for the phenomenon which were
translated properly, divided by the number of all test sentences for this
phenomenon.

This allows us also to perform comparisons among the systems, focusing on
particular phenomena.
The significance of every comparison between two systems is confirmed with a
two-tailed Z-test with $\alpha = 0.95$, testing the null hypothesis that the
difference between the two respective percentages is zero.

\subsection{Experiment setup}

The evaluation of the MT outputs was performed with TQ-AutoTest
\cite{MACKETANZ18.121}, a tool that organizes the test items in a database,
allowing the application of the regular expressions on new MT outputs.
For the purpose of this study, we have compared the 16 systems submitted to the
test suite task of the EMNLP2018 Conference of Machine Translation (WMT18) for
German$\rightarrow$English. At the time that this paper is written, the creators
of 11 of these systems have made their development characteristics available, 10
of them stating that they follow a NMT method and one of them a method combining
phrase-based SMT and NMT.

After the application of the existing regular expressions to the outputs of
these 16 systems, there was a considerable amount of warnings (i.e. uncertain
judgments) that varied between 10\% and 45\% per system.
A manual inspection of the outputs was consequently performed
(Figure~\ref{fig:construction}, stage~f) by a linguist, who invested
approximately 80 hours of manual annotation.
A small-scale manual inspection of the automatically assigned \textit{pass} and \textit{fail} labels indicated that the percentage of the erroneously assigned labels is negligible. 
The manual inspection therefore focused on warnings and reduced their amount to less than 10\%
warnings per system\footnote{Here, we do not take into account the two
unsupervised systems for the reasons explained in
Section~\ref{sec:results:systems}.}.
In particular, 32.1\% of the original system outputs ended in warnings, after
the application of the regular expressions, whereas the manual inspection and
the refining of the regular expressions additionally validated 14,000 of these
system outputs, i.e.
15.7\% of the original test suite.

In order to analyze the results with respect to the existence of warnings, we
performed two different types of analysis:
\begin{enumerate} 
\item Remove all sentences from the \textit{overall comparison} that have even one warning for one system and the translation accuracy on the remaining segments. The unsupervised systems are completely excluded from this analysis in order to keep the sample big enough. This way, all systems are compared on the same set of segments. 
\item Remove the sentences with warnings \textit{per system} and calculate the translation accuracy on the remaining segments. The unsupervised systems can be included in this analysis. In this way, the systems are \textit{not} compared on the same set of segments, but more segments can be included altogether. 
\end{enumerate}


 

\section{Results}

The final results of the evaluation can be seen in Table \ref{tab:ana1}, based
on Analysis~1 and Table~\ref{tab:ana2}, based on Analysis~2. 
Results for verb-related phenomena based on Analysis~1 are detailed in Tables~\ref{tab:tenses} and~\ref{tab:verbtypes} and other indicative phenomena in Table~\ref{tab:variousphenomena}.
The filtering prior to Analysis~1 left a small number of test sentences per
category, which limits the possibility to identify significant differences between the
systems.
Analysis~2 allows better testing of each system's performance, but observations
need to be treated with caution, since the systems are tested against different test sentences
and therefore the comparisons between them are not as expressive as in Analysis 1.
Moreover, the interpretability of the overall averages of these tables is
limited, as the distribution of the test sentences and the linguistic phenomena
does not represent an objective notion of quality.

We have calculated the mean values per system as non-weighted average and as weighted average. The non-weighted average was calculated by dividing the sum off all correct translations by the sum of all test sentences. The weighted average for a system was computed by taking a mean of the averages per category. We have not calculated statistical significances for the weighted averages as these are less meaningful due to the dominance of the verb tense/aspect/mood category. 

\subsection{Comparison between systems}\label{sec:results:systems}

The following results are based on Analysis 1. 
The system that achieves the highest accuracy in most linguistic phenomena, as
compared to the rest of the systems, is UCAM, which is in the first significance
cluster for 11 out of the 12 decisive error categories in Analysis 1 and
achieves a 86.0\% non-weighted average accuracy over all test sentences. 
UCAM obtains a significantly better performance than all other systems
concerning verb tense/aspect/mood, reaching a 86.9\% accuracy, 1.5\% better than
MLLP and NTT which are following in this category.
The different performance may be explained by the fact that UCAM differs
from others, since it combines several difference neural models together with a
phrase-based SMT system in an syntactic MBR-based scheme
\cite{DBLP:journals/corr/StahlbergGHB16}.
Despite its good performance in grammatical phenomena, UCAM has a very low
accuracy regarding punctuation (52.9\%).

The system with the highest weighted average score is RWTH. Even though it reaches higher accuracies for some categories than UCAM, the differences are not statistically significant. 

Another system that achieves the best accuracies at the 11 out of the 12
categories is Online-A.
This system performs close to the average of all systems concerning verb
tense/aspect/mood, but it shows a significantly better performance on the category
of punctuation (96.1\%).
Then, 6 systems (JHU, NTT, Online-B, Online-Y, RWTH, Ubiqus) have the best
performance at the same amount of categories (10 out of 12), having lost the
first position in punctuation and verb tense/aspect/mood.

Two systems that have the lowest accuracies in several categories are Online-F
and Online-G.
Online-F has severe problems with the punctuation (3.9\%) since it failed 
producing proper
quotation marks in the output and mistranslated other phenomena, such as commas
and the punctuation in direct speech (see Table~\ref{tab:variousphenomena}).
Online-G has the worst performance concerning verb tense/aspect/mood (45.8\%).
Additionally, these two systems together demonstrate the worst performance on
coordination/ellipsis and negation.

The \textbf{unsupervised systems} form a special category of systems trained
only on monolingual corpora.
Their outputs suffer from adequacy problems, often being very ``creative'' or
very far from a correct translation.
Thus, the automatic evaluation failed to check a vast amount of test sentences
on these systems. Therefore, we conducted Analysis 2. 
As seen in Table~\ref{tab:ana2}, unsupervised systems suffer mostly on MWE
(11.1\% - 17.4\% accuracy), function words (15.7\% - 21.7\%), ambiguity (26.9\% - 29.1\%) and non-verbal
agreement (38.3\% - 39.6\%).

\subsection{Linguistic categories}

Despite the significant progress in the MT quality, we managed to devise
test sentences that indicate that the submitted systems have a mediocre
performance for several linguistic categories.
On average, all current state-of-the-art systems suffer mostly on punctuation
(and particularly quotation marks), MWE, ambiguity and false friends with an
average accuracy of less than 64\% (based on Analysis 1).
Verb tense/aspect/mood, non-verbal agreement, function words and
coordination/ellipsis are also far from good, with average accuracies around
75\%. 

The two categories verb valency and named entities/terminology cannot lead to
comparisons on the performance of individual systems, since all systems achieve
equal or insignificantly different performance on them.
The former has an average accuracy of 81.4\%, while the latter has an average accuracy of 83.4\%.


We would like to present a few examples in order to provide a better
understanding of the linguistic categories and the evaluation.
Example (1) is taken from the category of \textbf{punctuation}.
Among others, we test the punctuation in the context of direct speech.
While in German it is introduced by a colon, in English it is introduced by a
comma.
In this example, the NTT system produces a correct output (therefore highlighted
in boldface), whereas the other two systems depict incorrect translations with a
colon.
\begin{table}[!h]
\begin{tabular}{p{.2\linewidth}p{.65\linewidth}r}
\multicolumn{2}{l}{(1) Punctuation} \\
\textit{source:}		& \textit{Er rief: \glqq Ich gewinne!\grqq} \\
\textbf{NTT:}		& He shouted, ``I win!" \\
Online-F:	& He called: ``I win!" \\
Ubiqus: 	& He cried: ``I win!" \\
\end{tabular}
\end{table}

\noindent We may assume that these errors are attributed to the fact that punctuation is often manipulated by hand-written pre- and post-processing tools, whereas the ability of the neural architecture to properly convey the punctuation sequence has attracted little attention and is rarely evaluated properly. 

\vspace{1.5px}
\noindent
\textbf{Negation} is one of the most important categories for meaning
preservation.
Two commercial systems (Online-F and Online-G) show the lowest accuracy for this
category and it is disappointing that they miss 4 out of 10 negations.
In Example (2), the German negation particle ``nie'' should be translated as
``never'', but Online-G omits the whole negation.
In other cases it negates the wrong element in the sentence.

\begin{table}[!h]
\begin{tabular}{p{.22\linewidth}p{.65\linewidth}r}
\multicolumn{2}{l}{(2) Negation}  \\
\textit{source:}	& \textit{Tim w\"ascht seine Kleidung nie selber.} \\
\textbf{Online-B:}	& Tim never washes his clothes himself.\\
Online-G:	& Tim is washing his clothes myself. \\
\end{tabular}
\end{table}

\noindent \textbf{MWE}, such as idioms or collocations, are prone to errors in
MT as they cannot be translated in their separate elements.
Instead, the meaning of the expression has to be translated as a whole.
Example~(3) focuses on the German idiom ``auf dem Holzweg sein'' which can be
translated as ``being on the wrong track".
However, a literal translation of ``Holzweg'' would be ``wood(en) way'',
``wood(en) track'' or ``wood(en) patch''.
As can be seen in the example, MLLP and UCAM provide a literal translation of
the separate segments of the MWE rather than translating the whole meaning of
it, resulting in a translation error.
	
\begin{table}[!h]
\begin{tabular}{p{.2\linewidth}p{.65\linewidth}r}
\multicolumn{2}{l}{(3) MWE}  \\
\textit{source:}	& \textit{Du bist auf dem Holzweg.} \\
MLLP:	& You're on the wood track. \\
\textbf{RWTH:}	& You're on the wrong track.  \\
UCAM:  	& You're on the wooden path.	\\
\end{tabular}
\end{table}

\subsection{Linguistic phenomena}

As mentioned above, a large part of the test suite is made up of verb-related phenomena. Therefore, we have conducted a more fine-grained analysis of the category ``Verb tense/aspect/mood''. In Table \ref{tab:tenses}  we have grouped the phenomena by verb tenses. Table \ref{tab:verbtypes} shows the results for the verb-related phenomena grouped by verb type.
Regarding the verb tenses, future II and future II subjunctive show the lowest accuracy with a maximum accuracy of about 30\%. The highest accuracy value on average (weighted and non-weighted) is achieved by UCAM with 63.5\%, respectively 61.5\%. UCAM is the only system that is one of the best-performing systems for all the verb tenses as well as for all the verb types. The second-best system on average for verb tenses and verb types is NTT. 

While the accuracy scores among the phenomena range between 33.4\% and 63.5\% for the verb tenses, the scores for the verb types are higher with 45.7\% - 86.9\%. 

Table \ref{tab:variousphenomena} shows interesting individual phenomena with at least 15 valid test sentences. The accuracy for compounds and location is generally quite high. There are other phenomena that exhibit a larger range of accuracy scores, as for example quotation marks, with an accuracy ranging from 0\% to 94.7\% among the systems. The system Online-F fails on all test sentences with quotation marks. The failure results from the system generating the quotation marks analogously to the German punctuation, e.g., introducing direct speech with a colon, as seen in Example (1). Online-F furthermore also fails on all test sentences with question tags, as does NJUNMT. For the phenomenon location, on the other hand, none of the systems is significantly better than any other system. They all perform similarly good, with an accuracy ranging from 86.7\% to 100\%. RWTH is the only system that reaches an accuracy of 100\% twice in these selected phenomena. 


\begin{table*}[ht] \linespread{0.6}\selectfont\centering
\setlength\extrarowheight{6pt}
\begin{tabular}{>{\raggedright\hangpara{10px}{1}}p{.23\linewidth}p{.65\linewidth}}
\toprule
category & phenomena \\
\midrule
Ambiguity & lexical ambiguity, structural ambiguity \\
Composition & phrasal verb, compound \\
Coordination \& ellipsis & slicing, right-node rasing, gapping, stripping \\
False friends & \\
Function word & focus particle, modal particle, question tag \\
 Long-distance dependency (LDD) \& interrogative & multiple connectors, topicalization, polar question,
WH-movement, scrambling, extended adjective construction, extraposition, pied-piping \\
Multi-word expression  & prepositional MWE, verbal MWE, idiom, collocation \\
Named entity (NE) \& terminology & date, measuring unit, location, proper name, domain-specific term \\
Negation & \\
Non-verbal agreement & coreference, internal possessor, external possessor \\
Punctuation & comma, quotation marks \\
Subordination & adverbial clause, indirect speech, cleft sentence, infinitive clause, relative clause, free relative clause, subject clause, object clause \\
Verb tense/aspect	& future I, future II, perfect, pluperfect, present, preterite, progressive \\
~~~~ mood   & indicative, imperative, subjunctive, conditional \\
~~~~ type   & ditransitive, transitive, intransitive, modal, reflexive   \\
Verb valency & case government, passive voice, mediopassive voice, resultative predicates \\
\bottomrule
\end{tabular}
\caption{Categorization of the grammatical phenomena}
\label{tab:categorization}
\end{table*}

\section{Conclusion and Further Work} 
We used a test suite in order to perform fine-grained evaluation in the output
of the state-of-the-art systems, submitted at the shared task of WMT18.
One system (UCAM), that uses a syntactic MBR combination of several NMT and
phrase-based SMT components, stands out regarding to verb-related phenomena.
Additionally, two systems fail to translate 4 out of 10 negations.
Generally, submitted systems suffer on punctuation (and particularly quotation
marks, with the exception of Online-A), MWE, ambiguity and false friends, and
also on translating the German future tense II.
6 systems have approximately the same performance in a big number of linguistic
categories.

Fine-grained evaluation would ideally provide the potential to identify particular flaws at the development of the translation systems and suggest specific modifications. 
Unfortunately, at the time that this paper was written, few details about the development characteristics of the respective systems were available, so we could provide only few assumptions based on our findings. 
The differences observed may be attributed to the design of the models, to pre- and post-processing tools, to the amount, the type and the filtering of the corpora and other development decisions.
We believe that the findings are still useful for the original developers of the systems, since they are aware of all their technical decisions and they have the technical possibility to better inspect the causes of specific errors.

\section*{Acknowledgments} 
This work was supported by XXX through the project Open Source Lab and by the German Federal Ministry of Education and Research (BMBF) through the project DEEPLEE (01lW17001).



Special thanks to Arle Lommel and Kim Harris who helped with their input in
earlier stages of the experiment, to Renlong Ai and He Wang who developed
and maintained the technical infrastructure and to Aylin Cornelius who helped with the evaluation.

\bibliographystyle{acl_natbib_nourl} 
\bibliography{all}


\begin{sidewaystable}
\small  
\centering      
\begin{tabular}{lrcccccccccccccc}
\toprule
                             & \#   & JHU              & LMU         & MLLP         & NJUNMT  		  & NTT              & onl-A        & onl-B         & onl-F         & onl-G         & onl-Y      	& RWTH           & Ubiqus        & UCAM           & uedin           \\
\midrule
Ambiguity                    & 76   & \textbf{69.7}  & \textbf{64.5} & 55.3           & 59.2          & \textbf{63.2}  & \textbf{68.4} & \textbf{73.7}  & 42.1          & \textbf{71.1} & \textbf{65.8} & \textbf{78.9}  & \textbf{64.5} & \textbf{76.3}  & 51.3          \\
False friends                & 34   & \textbf{61.8}  & \textbf{58.8} & \textbf{61.8}  & 50.0          & \textbf{73.5}  & \textbf{76.5} & \textbf{79.4}  & \textbf{70.6} & \textbf{76.5} & \textbf{67.6}	& \textbf{70.6}  & \textbf{55.9} & \textbf{67.6}  & \textbf{55.9} \\
Verb valency                 & 30   & 80.0           & 73.3          & 86.7           & 83.3          & 86.7           & 86.7          & 86.7           & 70.0          & 76.7          & 86.7          & 90.0           & 86.7          & 86.7           & 86.7          \\
Verb tense/aspect/mood       & 4110 & 74.3           & 65.8          & 84.4           & 61.6          & 84.3           & 71.8          & 72.1           & 75.2          & 45.8          & 70.6          & 78.3           & 73.5          & \textbf{86.9}  & 76.7          \\
Non-verbal agreement         & 48   & \textbf{75.0}  & 60.4          & \textbf{79.2}  & \textbf{77.1} & \textbf{83.3}  & \textbf{75.0} & \textbf{87.5}  & 50.0          & 60.4          & \textbf{81.3} & \textbf{85.4}  & \textbf{75.0} & \textbf{81.3}  & \textbf{75.0} \\
Punctuation                  & 51   & 60.8           & 56.9          & 62.7           & 56.9          & 68.6           & \textbf{96.1} & 74.5           & 3.9           & 60.8          & 76.5          & 58.8           & 82.4          & 52.9           & 66.7          \\
Subordination                & 34   & \textbf{94.3}  & \textbf{94.3} & \textbf{91.4}  & \textbf{91.4}& \textbf{94.3}   & \textbf{94.3} & \textbf{85.7}  & 45.7	        & \textbf{77.1} & \textbf{94.3} & \textbf{88.6}  & \textbf{88.6} & \textbf{91.4}  &  \textbf{91.4}        \\
MWE                          & 54   & \textbf{63.0}  & \textbf{55.6} & \textbf{59.3}  & \textbf{66.7} & \textbf{66.7}  & \textbf{68.5} & \textbf{75.9}  & 42.6          & \textbf{70.4} & \textbf{75.9} & \textbf{70.4}  & \textbf{61.1} & \textbf{66.7}  & \textbf{61.1} \\
LDD \& interrogatives        & 40   & \textbf{80.0}  & \textbf{77.5} & \textbf{80.0}  & \textbf{88.0} & \textbf{82.5}  & \textbf{82.5} & \textbf{85.0}  & 60.0          & \textbf{77.5} & \textbf{85.0}  & \textbf{87.5} & \textbf{82.5} & \textbf{87.5}  & \textbf{75.0}         \\
NE \& terminology 			 & 35   & 82.9           & 80.0          & 88.6           & 80.0          & 88.6           & 77.1          & 77.1           & 77.1          & 77.1          & 91.4          & 82.9           & 77.1          & 80.0           & 80.0          \\
Coordination \& ellipsis     & 24   & \textbf{79.2}  & \textbf{79.2} & \textbf{87.5}  & \textbf{79.2} & \textbf{87.5}  & \textbf{87.5} & \textbf{87.5}  & 25.0          & \textbf{58.3} & \textbf{87.5} & \textbf{87.5}  & \textbf{87.5} & \textbf{87.5}  & \textbf{83.3} \\
Negation					 & 20	& \textbf{100.0} & \textbf{90.0} & \textbf{100.0} & \textbf{95.0 } & \textbf{100.0} & \textbf{100.0}& \textbf{95.0}	& 65.0		    & 60.0			& \textbf{100.0}& \textbf{100.0} & \textbf{95.0} & \textbf{100.0} & \textbf{100.0} 		\\
Composition                  & 43   & \textbf{83.7}  & 60.5          & \textbf{81.4} & 69.8 		  & \textbf{88.4}  & \textbf{79.1} & \textbf{95.3}  & \textbf{76.7} & 72.1          & \textbf{88.4} & \textbf{88.4}  & \textbf{76.7} & \textbf{93.0}  & \textbf{76.7} \\
Function word                & 50   & \textbf{72.0}  & 56.0          & \textbf{76.0}  & 52.0          & \textbf{76.0}  & \textbf{82.0} & \textbf{76.0}  & 38.0          & 48.0          & \textbf{86.0} & \textbf{88.0}  & \textbf{78.0} & \textbf{80.0}  & \textbf{78.0} \\
\midrule
Sum                          & 4650 & 3463           & 3071         & 3873            & 2913          & 3893           &3393           & 3413           & 3379          & 2262          & 3344          & 3663           & 3438          & 4005           & 3547            \\
Non-weighted average         &      & 74.4           & 66.0          & 83.3           & 62.6          & 83.7           & 72.9          & 73.3           & 72.6          & 48.5          & 71.9          & 78.7           & 73.8          & \textbf{86.0}  & 76.2   \\
Weighted average 			 &		& 76.9		   	 & 69.5			& 78.2			 & 71.6			& 81.7				& 81.8			& 82.2			& 53.0			& 66.6			& 82.6			& 82.5			& 77.5			& 81.3				& 75.6	\\
\bottomrule
\end{tabular}
		\caption{System accuracy (\%) on each error category based on Analysis 1,
		having removed all test sentences whose evaluation remained uncertain, even
		for one of the systems. Boldface indicates the significantly best systems in the
		category}
		\label{tab:ana1}
\end{sidewaystable}

\begin{sidewaystable}
\small
\centering      
\begin{tabular}{lcccccccccccccccc}
\toprule
                             &JHU    & LMU 	  & LMU-uns& MLLP 		& NJUNMT& NTT   & onl-A & onl-B & onl-F & onl-G & onl-Y 	& RWTH-uns& RWTH  & Ubiqus& UCAM & uedin           \\
\midrule
Ambiguity                    & 65.4  & 61.3   & 26.9  & 53.1        & 59.5  & 59.3  & 66.7  & 74.7  & 43.2  & 70.4  & 67.1      & 29.1  & 80.3    & 63.6  & 72.4  & 48.2  \\
False friends                & 58.3  & 55.6   & 50.0  & 58.3        & 47.2  & 73.5  & 72.2  & 75.0  & 66.7  & 72.2  & 66.7      & 41.7  & 66.7    & 52.8  & 63.9  & 52.8  \\
Verb valency                 & 76.7  & 67.2   & 60.0  & 84.5        & 76.8  & 86.4  & 81.4  & 84.1  & 55.3  & 67.9  & 89.5      & 50.0  & 91.7    & 78.7  & 80.7  & 74.2  \\
Verb tense/aspect/mood       & 73.4  & 64.6   & 29.4  & 83.0        & 61.0  & 83.7  & 71.5  & 71.9  & 74.3  & 44.9  & 70.4      & 42.5  & 77.6    & 72.2  & 85.7  & 75.8  \\
Non-verbal agreement         & 77.4  & 56.1   & 38.3  & 81.5        & 75.9  & 85.5  & 77.6  & 89.7  & 46.4  & 59.7  & 79.3      & 39.6  & 86.0    & 72.9  & 82.5  & 75.0  \\
Punctuation                  & 63.8  & 59.6   & 50.0  & 65.5        & 60.7  & 70.2  & 96.5  & 75.4  & 3.5   & 61.1  & 76.8      & 51.0  & 61.4    & 84.2  & 56.9  & 69.0  \\
Subordination                & 98.8  & 98.5   & 79.2  & 96.4        & 97.6  & 97.8  & 98.8  & 95.7  & 70.5  & 89.7  & 98.9      & 75.0  & 96.7    & 96.3  & 96.4  & 97.3 \\
MWE                          & 63.1  & 49.3   & 11.1  & 56.3        & 62.7  & 62.9  & 63.8  & 70.8  & 40.9  & 65.7  & 66.2      & 17.4  & 67.1    & 58.6  & 64.8  & 57.4  \\
LDD \& interrogatives        & 83.7  & 72.6   & 73.3  & 86.4        & 81.8  & 86.8  & 83.1  & 88.3  & 53.6  & 56.2  & 86.0      & 81.3  & 93.3    & 85.4  & 88.7  & 76.6  \\
NE \& terminology            & 86.5  & 79.7   & 84.4  & 86.5        & 83.3  & 87.7  & 78.4  & 81.8  & 75.0  & 72.6  & 84.3      & 91.7  & 87.0    & 77.8  & 78.9  & 79.5  \\
Coordination \& ellipsis     & 83.1  & 80.0   & 61.5  & 84.1        & 78.0  & 88.9  & 86.7  & 92.9  & 26.1  & 55.1  & 80.0      & 18.5  & 88.0    & 84.1  & 90.7  & 86.5  \\
Negation                     & 100.0 & 90.0   & 23.5  & 100.0       & 95.0  & 100.0 & 100.0 & 95.0  & 65.0  & 60.0  & 100.0     & 57.1  & 100.0   & 95.0  & 100.0 & 100.0 \\
Composition                  & 81.3  & 59.2   & 60.0  & 79.2        & 67.3  & 85.4  & 77.6  & 93.6  & 76.1  & 70.8  & 87.5      & 14.6  & 85.7    & 73.5  & 89.8  & 73.5  \\
Function word                & 70.4  & 51.4   & 15.7  & 78.3        & 50.7  & 76.4  & 81.2  & 75.3  & 32.8  & 47.1  & 84.7      & 21.7  & 81.9    & 71.2  & 80.6  & 75.3  \\
\midrule
Sum                          & 5328  & 5287   & 1795  & 5303        & 5231  & 5331  & 5312  & 5357  & 5211  & 5171  & 5309      & 2341  & 5362    & 5296  & 5351  & 5300     \\
Non-weighted average         & 74.1  & 65.0   & 31.8  & 82.2        & 62.9  & 83.3  & 73.1  & 73.9  & 70.6  & 47.9  & 72.3      & 41.6  & 78.8    & 73.0  & 84.9  & 75.5 \\
Weighted average			 & 77.3	 & 67.5	  & 47.4  & 78.1		& 71.3	& 81.7	& 81.1	& 83.2	& 52.1	& 63.8	& 81.2		& 45.1	& 83.1	  & 76.2  & 80.9  & 74.4 \\
\bottomrule
\end{tabular}
		\caption{System accuracy (\%) on each error category based on Analysis 2, having removed only the system outputs whose evaluation remained uncertain.}
		\label{tab:ana2}
\end{sidewaystable}


\begin{sidewaystable}
\small
\centering
\begin{tabular}{lrcccccccccccccc}
\toprule
                          & \#  & JHU  & LMU  & MLLP &NJUNMT& NTT  & onl-A& onl-B& onl-F& onl-G& onl-Y& RWTH &Ubiqus& UCAM & uedin\\
\midrule
Future I                  & 494 & \textbf{70.2} & 68.3 & \textbf{70.3} & 56.8 & \textbf{69.7} & 64.8 & 65.7 & 64.8 & 45.8 & 63.8 & 68.8 & 68.5 & \textbf{73.8} & \textbf{69.3} \\
Future I subjunctive II   & 479 & 66.8 & 56.3 & \textbf{75.7} & 48.3 & \textbf{74.5} & 63.8 & 60.5 & 44.2 & 39.8 & 59.3 & 64.7 & 63.7 & \textbf{73.8} & 66.7 \\
Future II                 & 138 & \textbf{28.8} & \textbf{27.5} & \textbf{28.5} & \textbf{23.8} & \textbf{27.8} & 21.0 & \textbf{31.0} & \textbf{25.3} & 12.0 & \textbf{25.8} & \textbf{30.3} & \textbf{28.5} & \textbf{30.8} & \textbf{29.0} \\
Future II subjunctive II  & 128 & \textbf{27.5} & 17.0 & \textbf{27.5} & \textbf{23.3} & \textbf{29.5} & \textbf{29.3} & \textbf{31.0} & \textbf{26.3} & 18.0 & 30.0 & \textbf{26.5} & 11.3 & \textbf{30.3} & \textbf{25.3} \\
Perfect                   & 506 & 64.8 & 51.2 & \textbf{80.5} & 60.0 & \textbf{79.7} & 57.0 & 65.5 & 67.7 & 26.5 & 67.7 & 74.8 & 62.8 & \textbf{78.0} & 70.3 \\
Pluperfect                & 478 & 39.7 & 24.7 & \textbf{57.0} & 23.5 & \textbf{52.5} & 22.3 & 26.5 & \textbf{54.2} & 16.0 &  5.0 & 30.2 & 33.7 & \textbf{52.3} & 50.2 \\
Pluperfect subjunctive II & 442 & 37.2 & 35.8 & \textbf{53.0} & 37.3 & \textbf{54.2} & 41.0 & 40.7 & 50.3 & 13.7 & 49.5 & \textbf{52.2} & 42.7 & \textbf{56.3} & 41.3 \\
Present                   & 482 & 71.3 & 68.3 & \textbf{73.5} & 58.5 & 69.7 & 69.8 & 59.8 & 68.8 & 50.2 & 61.8 & \textbf{75.5} & 72.0 & \textbf{77.7} & \textbf{75.0} \\
Preterite                 & 513 & 69.3 & 66.2 & 74.3 & 59.3 & 79.5 & \textbf{81.2} & 78.0 & 70.7 & 60.2 & \textbf{80.7} & 76.2 & 76.3 & \textbf{83.7} & 64.3 \\
Preterite subjunctive II  & 433 & 49.8 & 48.0 & \textbf{54.2} & 44.2 & \textbf{57.2} & \textbf{56.0} & \textbf{53.5} & \textbf{58.3} & 39.3 & \textbf{56.0} & \textbf{53.5} & \textbf{55.0} & \textbf{56.2} & 49.3 \\
\midrule
Sum/non-weighted average             	 & 4093 & 54.3 & 48.1 & 61.7 & 44.9 & 61.6 & 52.4 & 52.7 & 55.0 & 33.4 & 51.5 & 57.2 & 53.7 & 63.5 & 56.0 \\
Weighted average 		 &		&52.7  & 46.5 & 59.7 & 43.6	& 59.6 & 50.8 & 51.4 & 53.2 & 32.3 & 50.1 & 55.4 & 51.7 & 61.5 & 54.2	\\
\bottomrule
\end{tabular} 
\caption{System accuracy (\%) on linguistic phenomena related to verb tenses}
\label{tab:tenses}
\end{sidewaystable}
\begin{sidewaystable}
\small

\centering

\begin{tabular}{lccccccccccccccc}

\toprule
                & \#  & JHU  & LMU  & MLLP &NJUNMT& NTT  & onl-A& onl-B& onl-F& onl-G& onl-Y& RWTH &Ubiqus& UCAM & uedin\\

\midrule
Ditransitive    & 329 & 85.7 & 82.7 & 90.0 & 70.8 & 93.6 & 77.8 & 88.8 & 79.0 & 61.1 & 81.5 & 90.0 & 79.0 & \textbf{95.4} & 83.9 \\
Intransitive    & 397 & 77.1 & 77.6 & 89.2 & 64.5 & 84.6 & 83.9 & 89.9 & 78.1 & 69.3 & 87.7 & 90.4 & 84.1 & \textbf{93.7} & 86.6 \\
Modal          & 1353 & 69.0 & 56.4 & \textbf{82.6} & 56.2 & \textbf{81.5} & 65.9 & 68.1 & 73.7 & 39.2 & 65.0 & 73.1 & 69.3 & \textbf{81.7} & 71.8 \\
Modal negated  & 1403 & 73.5 & 65.6 & \textbf{85.3} & 59.2 & 82.2 & 66.6 & 59.0 & 74.8 & 36.4 & 63.2 & 74.1 & 71.7 & \textbf{86.2} & 75.2 \\
Reflexive       & 246 & 74.0 & 50.8 & 58.9 & 45.1 & \textbf{80.9} & 75.6 & \textbf{85.4} & 64.6 & 30.9 & 73.2 & 69.9 & 61.8 & \textbf{80.1} & 71.5 \\
Transitive      & 365 & 83.6 & 82.7 & 94.8 & 89.0 & \textbf{96.2} & 92.1 & 93.2 & 83.6 & 75.6 & 88.8 & 95.1 & 86.8 & \textbf{98.1} & 85.8 \\
\midrule
Sum/non-weighted average    & 4093 & 74.3 & 65.7 & 84.4 & 61.5 & 84.3 & 71.8 & 72.0 & 75.3 & 45.7 & 70.5 & 78.2 & 73.5 & 86.9 & 76.6 \\
Weighted average &	  & 77.2 & 69.3 & 83.5 & 64.1 & 86.5 & 77.0 & 80.7 & 75.6 & 52.1 & 76.6 & 82.1 & 75.5 & 89.2 & 79.1 	\\
\bottomrule
\end{tabular} 
\caption{System accuracy (\%) on linguistic phenomena related to verb types}
\label{tab:verbtypes}
\end{sidewaystable}


\begin{sidewaystable}
\small
\centering
\begin{tabular}{lrcccccccccccccc}
\toprule
                & \# & JHU   & LMU  & MLLP & NJUNMT & NTT   & onl-A & onl-B & onl-F & onl-G & onl-Y & RWTH  & Ubiqus & UCAM  & uedin   \\
\midrule
Compound        & 26 & 73.1  & 73.1 & 73.1 & 69.2   & \textbf{84.6}  & \textbf{80.8}  & \textbf{92.3}  & \textbf{76.9}  & \textbf{80.8}  & \textbf{84.6}  & \textbf{80.8}  & \textbf{76.9}   & \textbf{88.5}  & \textbf{76.9}    \\
Quotation marks & 38 & 47.4  & 42.1 & 50.0 & 42.1   & 60.5  & \textbf{94.7}  & 68.4  & 0.0   & 50.0  & 68.4  & 44.7  & 76.3   & 36.8  & 55.3    \\
Phrasal verb    & 17 & \textbf{100.0} & 58.8 & \textbf{94.1} & 70.6   & \textbf{94.1}  & 76.5  & \textbf{100.0} & 76.5  & 58.8  & \textbf{94.1}  & \textbf{100.0} & 76.5   & \textbf{100.0} & 76.5    \\
Question tag    & 15 & 66.7  & 20.0 & \textbf{86.7} & 0.0    & 73.3  & 66.7  & 60.0  & 0.0   & 13.3  & \textbf{93.3}  & \textbf{100.0} & 73.3   & \textbf{86.7}  & \textbf{80.0}    \\
Collocation     & 15 & 60.0  & 40.0 & 53.3 & 60.0   & 60.0  & \textbf{66.7}  & \textbf{86.7}  & 20.0  & 80.0  & \textbf{86.7}  & 60.0  & 60.0   & \textbf{66.7}  & 60.0    \\
Location        & 15 & 93.3 & 86.7 & 93.3 & 86.7 & 100.0 & 86.7 & 86.7 & 93.3 & 93.3 & 100.0 & 93.3 & 93.3 & 93.3 & 86.7 \\
Modal particle  & 16 & 56.3  & 50.0 & 56.3 & 50.0   & \textbf{75.0}  & \textbf{93.8}  & \textbf{81.3}  & 18.8  & 50.0  & \textbf{87.5}  & \textbf{75.0}  & 56.3   & 62.5  & 56.3    \\
\bottomrule
\end{tabular}
\caption{System accuracy (\%) on specific linguistic phenomena with more than 15
test sentences}
\label{tab:variousphenomena}
\end{sidewaystable}



\end{document}